# Extraction of Visual Information to Predict Crowdfunding Success

Simon J. Blanchard
McDonough School of Business, Georgetown University, Washington, DC, 20057, USA
sjb247@georgetown.edu

Theodore J. Noseworthy
Schulich School of Business, York University, North York, ON, M3J 1P3, Canada
tnoseworthy@schulich.yorku.ca

Ethan Pancer
Sobey School of Business, Saint Mary's University, Halifax, NS, B3H 3C3, Canada
ethan.pancer@smu.ca

Maxwell Poole
Sobey School of Business, Saint Mary's University, Halifax, NS, B3H 3C3, Canada
maxwellpoole@gmail.com

* Contact author: Simon J. Blanchard, Georgetown University, 37th and O St. N.W., Washington, D.C. 20057. Phone: 202-687-6977. E-mail: sjb247@georgetown.edu. The authors would like to thank Zhengrong Gu for technical assistance, and seminar participants at UC San Diego, Fudan University, McGill University, Shanghai Tech, and the Theory and Practice in Marketing Conference for helpful feedback.

# Extraction of Visual Information to Predict Crowdfunding Success

Researchers have increasingly turned to crowdfunding platforms to gain insights into entrepreneurial activity and dynamics. While previous studies have explored various factors influencing crowdfunding success, such as technology, communication, and marketing strategies, the role of visual elements that can be automatically extracted from images has received less attention. This is surprising, considering that crowdfunding platforms emphasize the importance of attention-grabbing and high-resolution images, and previous research has shown that image characteristics can significantly impact product evaluations. Indeed, a comprehensive review of empirical articles (n = 202) that utilized Kickstarter data, focusing on the incorporation of visual information in their analyses. Our findings reveal that only 29.70% controlled for the number of images, and less than 12% considered any image details.

      In this manuscript, we contribute to the existing literature by emphasizing the significance of visual characteristics as essential variables in empirical investigations of crowdfunding success. We review the literature on image processing and its relevance to the business domain, highlighting two types of visual variables: visual counts (number of pictures and number of videos) and image details. Building upon previous work that discussed the role of color, composition and figure-ground relationships, we introduce visual scene elements that have not yet been explored in crowdfunding, including the number of faces, the number of concepts depicted, and the ease of identifying those concepts.

      To demonstrate the predictive value of visual counts and image details, we analyze Kickstarter data using flexible machine learning models (Lasso, Ridge, Bayesian Additive Regression Trees, and XGBoost). Our results highlight that visual count features are two of the top three predictors of success and highlight the ease at which researchers can incorporate some information about visual information. Our results also show that simple image detail features such as color matter a lot, and our proposed measures of visual scene elements can also be useful. By supplementing our article with R and Python codes that help authors extract image details (https://osf.io/ujnzp/), we hope to stimulate scholars in various disciplines to consider visual information data in their empirical research and enhance the impact of visual cues on crowdfunding success.

**Keywords**: image characteristics, visual scene processing, new ventures, crowdfunding

# 1. INTRODUCTION

Crowdfunding allows entrepreneurs to request funding from many investors simultaneously, usually in exchange for future products or equity stakes (Mollick 2014; Mollick and Nanda 2015). Recent legislative changes also give consumers and firms new ways to interact along these lines. Since 2016, when the U.S. Securities and Exchange Commission (SEC) allowed investors of various net worths to participate in crowdfunding, there has been a substantial increase in the amount of money raised on crowdfunding platforms; in 2020, it reached $500 million in the United States alone, according to the Crowdfunding Center.[1] Dozens of different crowdfunding platforms support rewards-based investing (e.g., Kickstarter, Indiegogo) and equity-based investing (e.g., SeedInvest, StartEngine), and others specialize in individual help (e.g., GoFundMe) and nonprofit funding (e.g., Mightycause). The number of international crowdfunding platforms also has grown, with platforms covering all regions of the world (e.g., JD Crowdfunding and Jingdong in China, Babyloan in France, Crowdprop in South Africa).

Given the growth of crowdfunding as an alternative funding source in recent years, researchers have increasingly turned to datasets from various crowdfunding platforms to gain insights into entrepreneurial activity and the dynamics of crowdfunding. Some operations management scholars have leveraged crowdfunding data to substantiate the role of new technologies in selecting and disseminating new ideas (Kornish and Hutchinson-Krupat 2017), while others have sought to identify predictors of crowdfunding success (e.g., Cui, Kumar, and Gonçalves 2019; Wei et al. 2021), such as backer dynamics (Kuppuswamy and Bayus 2018), referral timing (Burtch, Gupta, and Martin 2021), founder updates (Mejia, Urrea, and Pedraza-Martinez 2019), and frequency of communication (Xiao, Ho, and Che 2021). Marketing scholars have typically focused on predictors of successful communication within project pages, such as information in videos (Li, Shi, and Wang 2019), emotions in text (Peng et al. 2021), and price advertising claims (Blaseg, Schulze, and Skiera 2020).

Recently, more attention is being focused on the role of visual elements in predicting crowdfunding success. This interest is likely due to three factors. First, crowdfunding platforms themselves provide guidance on how to develop project images that are most likely to attract backers. For instance, Kickstarter recommends that main project images be attention-grabbing, straightforward, and of high resolution[2], while Indiegogo stresses that image style and content are critical factors for success[3]. This guidance indicates that platforms recognize the significant role played by images in a project's success. Second, previous research has established that image characteristics such as brightness and contrast, as well as image compositions, can influence the persuasiveness of new product depictions. For example, studies by Noseworthy and Trudel (2011) and Pieters, Wedel, and Batra (2010) have found that altering image characteristics can significantly impact product evaluations. Third, empirical researchers have increasingly been using vision APIs and algorithms to augment their datasets and investigate the role of visual figures on company success (e.g., Zhang and Luo 2023, in the context of Yelp, Zhang et al., 2022, in the context of AirBnB).

---

[1] https://www.thecrowdfundingcenter.com/
[2] https://www.kickstarter.com/help/images
[3] https://support.indiegogo.com/hc/en-us/articles/206447157-Make-Your-Story-Captivating-with-Images-Videos

But to what extent do empirical researchers interested in crowdfunding utilize visual information in their analyses? As Kickstarter datasets are among the most frequently used sources of crowdfunding data due to their availability from sources like webrobots and Kaggle, and they serve as the foundation for our empirical study, we compiled a list of 399 empirical papers in operations, marketing, and management that utilized Kickstarter data. Our aim was to quantify the extent to which these articles incorporated visual information in their analyses, focusing on two types: visual counts (number of images and number of videos) and image details. After conducting a thorough manual review of the papers to include empirical analyses relevant to Kickstarter, [4] we identified 202 articles for which we could assess whether variables included counts of images (either 0 or 1+ or a count), counts of videos (either 0 or 1+ or a count), and *any* image detail (e.g., colorfulness, contrast). The results are presented in Table 1, and the corresponding spreadsheet is available on OSF.

Table 1 – Use of visual in empirical analyses of Kickstarter datasets

|  | **Number of Articles** | **Visual Count: Images** | **Visual Count: Videos** | **Any Image Detail** |
| --- | --- | --- | --- | --- |
| Management | 146 | 27.40% | 43.84% | 9.59% |
| Marketing | 26 | 46.15% | 50.00% | 30.77% |
| Operations Management | 30 | 26.67% | 50.00% | 6.67% |
| **Total** | **202** | **29.70%** | **45.54%** | **11.88%** |

Our findings indicate that approximately 46% of the empirical articles included a control variable to account for the presence of videos. However, only 29.70% of the articles controlled for the number of images, and less than 12% incorporated *any* image details in their analyses. Notably, we observed significant differences between subfields regarding the inclusion of visual characteristics. While management, marketing, and operations articles were roughly equally likely to have a variable capturing the presence of one (or more) videos, marketing articles were nearly twice as likely to include a variable for the count of images and were also more likely to consider image details. These highlight an opportunity to investigate potential important predictors insufficiently considered.

---

[4] First, we created a list of journals from the ABDC journal lists for three fields of research: Marketing; Strategy,Management, and Organizational Behavior; and Management Science, Operations Research, and Logistics. This list included a total of 615 journals. Second, we searched for articles in those journals containing the keyword "Kickstarter" anywhere with at least one of the words "webrobots" OR "kaggle" using Google Scholar. After manual cleaning for duplicates, we were left with 358 articles. Third, we added 41 articles which had a phrase match of "Kickstarter dataset."

In the present manuscript, our aim is to contribute to the existing literature on image processing and crowdfunding by emphasizing the significance of visual characteristics as essential variables in empirical investigations of the factors that contribute to crowdfunding success. To accomplish this, we undertake the following approach. First, we conduct a comprehensive review of the literature on image processing and its relevance to the business domain. This review highlights the importance of two types of visual variables that should not be overlooked: visual counts and image details. Second, we expand upon the work of Zhang and Luo (2023) by incorporating their typology of image details, which encompasses factors such as color, composition, and figure-ground. In addition to these aspects, we introduce visual scene elements that have not yet been explored in the context of crowdfunding. These scene elements include variables such as the number of faces present, the number of concepts depicted, and the ease of identifying those concepts. Finally, we present an analysis of Kickstarter data to demonstrate the predictive value of visual counts and image details. We employ nested variable sets across flexible models, including Lasso, Ridge, Bayesian Additive Regression Trees, and eXtreme Gradient Boosted Trees. We find consistent improvement when adding visual variables, particularly visual count variables, which significantly enhance both in-sample and out-of-sample RMSE and MAE metrics. Notably, these visual counts influence predictions nearly as much as the entire LIWC dictionary for text. We also find that BART and XGBoost as implemented consistently outperform Ridge and Lasso, suggesting better accommodation of non-linearities using untransformed data. While image details slightly improve out-of-sample performance over baseline variables, their predictive value is less than that of visual count variables.

Our work makes two main contributions. First, we have observed that, despite the increasing empirical analyses of crowdfunding data, there is still a lack of attention given to image-related features. By emphasizing the importance and relative ease of controlling for visual counts and image details, we hope to encourage empirical researchers to include visual counts and image details in their studies of crowdfunding as they are important predictors of crowdfunding success. By supplementing our work with specific implementation and measures, we seek to stimulate crowdfunding, marketing, operations, and entrepreneurship scholars to expand their consideration of visual information data in their empirical research.

Second, much recent research has focused on automated feature creation can be used to take advantage of multi-media formats such as video and images (Li, Shi and Wang 2019; Zhang and Luo 2023; Zhang et al. 2022). Most of that literature on automated feature extraction has been focused on visual processing characteristics (e.g., color, composition and figure-ground) that are related more to execution than meaning. In addition to introducing several new such features (blur, number of segments, image quality/noise), we highlight the usefulness and ease of using APIs to automatically extract visual-scene elements such as the number of faces, the number of evoked concepts, and the ease of concept identification. Our research box contains the list of 399 articles we surveyed, the Python codes for our visual feature extractions, the final dataset for analyses, analysis codes in R are available at https://osf.io/ujnzp/. We also offer a detailed guide at https://bit.ly/ujnzp. By sharing these files, we aim to facilitate the understanding of the impact of visual cues on crowdfunding success and across disciplines.

## 2. A TAXONOMY OF VISUAL INFORMATION

A substantial body of literature suggests that specific image characteristics can have a significant impact on human judgment and decision-making processes across various contexts. Previous research on visual information has predominantly focused on two aspects: accounting for visual count variables, such as the number of pictures and videos, and examining the photographic properties of an image that are relevant to perception and interpretation. These properties include color features, composition, and figure-ground (Zhang & Luo, 2023; Zhang et al., 2022). In this section, we provide a brief summary of these image characteristics and explain why we anticipate their influence on the evaluation of new products in the context of crowdfunding. Additionally, we discuss the potential utility of considering visual scene elements in crowdfunding analysis.

2.1. Visual Count Variables: Number of pictures and number of videos

According to Kickstarter's Creator Handbook (2022), images and videos play a crucial role in engaging potential backers and conveying the essence of a project. They serve as the initial points of contact, capturing backers' attention and making a favorable first impression. Scholarly research has recognized the significance of visual information in empirical studies (Blaseg et al., 2020; Mejia et al., 2019; Li et al., 2020). Mollick (2014) argues that incorporating videos indicates better preparation and higher project quality, which, in turn, leads to greater crowdfunding success (see also Kuppuswamy and Bayus, 2018; Tian et al., 2021; Wei et al., 2021). The number of images featured in projects also has a positive influence on backer contributions (Wei et al., 2021; Xiao et al., 2021). Project images offer a depth of communication that surpasses textual information, such as showcasing prototypes, and enhance perceptions of campaign informativeness, creator commitment, and credibility (Xiao et al., 2021).

Mollick and Nanda (2016) suggest that crowds, as opposed to experts, prioritize visual content, making projects with more images and videos more appealing to the crowd. Moreover, the literature on information processing suggests that increasing the number of informational cues moderately enhances persuasion (e.g., Brown and Carpenter, 2000). Therefore, it is crucial to include variables that capture visual information, such as the number of pictures and videos. However, our findings indicate that the majority of projects do not incorporate visual counts in their analyses.

2.2. Image Details

*2.2.1 Color features, composition and figure-ground relationship*

In addition to considering the numbers of pictures and videos in crowdfunding projects, it is important to examine the characteristics of the images themselves as they can impact decision-making. The field of color psychology has provided scientific insights into how perceptions of color influence attitudes and choices, with color carrying meaning and influencing affect, cognition, and behavior in consumer contexts (Elliot and Maier, 2014). Practitioners in marketing, advertising, and graphic design have long recognized that color and composition play a role in influencing consumer behavior through factors such as atmospherics (Gorn et al., 2004), brand identity (Labrecque and Milne, 2013), and product evaluations (Miller and Kahn, 2005).Common applications of computer vision algorithms have made the automatic extraction of *color* features such as brightness (amount of light reflected), saturation (color intensity), colorfulness (chromaticity), contrast (variation in luminance), warmth hue (degree of red, orange and yellow) and clarity (sharpness) relatively easy (see Zhang and Luo 2023). Most color feature characteristics can be calculated directly from pixel values using pre-determined rules and transformations (see Table 2). Others, like image quality (absence of noise), is usually done by trained classifiers (Mittal et al. 2012).

Image *composition* refers to the arrangement of visual elements within an image. Similar to color features, there are various ways to automatically extract variables that characterize image composition. Zhang and Luo (2023) focused on several composition characteristics, including diagonal dominance (average distance of salient points to the diagonal), the use of the rule of thirds (placement of elements at intersecting lines dividing the image into equal ninths), physical balance (distribution of elements across horizontal and vertical lines), and color balance (even distribution of colors across the image). Other composition characteristics include blur (difficulty in perceiving shapes) and the number of image segments. Algorithms such as OpenCV and the superpixel method developed by Anchanta et al. (2010) can be used to extract these composition variables by identifying salient regions and dividing the image into a grid.

Similarly, pre-trained algorithms can be employed to identify figure-ground relationships, which involve distinguishing the main object from its surrounding. Common figure-ground relationship variables include size difference (difference in area size between foreground and background), color difference (contrast between the foreground and background colors), and texture (variations in edge pixels between the foreground and background). Automatic extraction of figure-ground relationship variables typically involves using a classifier to separate the foreground from the background (e.g., GrabCut by Rother, Kolmogorov, and Blake, 2001) before calculating the differences. Each of these image characteristics, including color features, composition, and figure-ground relationships, provides information about the quality and content of the images. They help identify blurry images or those of lower quality. However, other types of distortions can also impact image quality, such as noise, compression artifacts, transmission errors, and intensity shifts (Ponomarenko et al., 2009). Computer scientists have developed algorithms that approximate human judgments of overall image quality and noise perception, providing a more holistic assessment (Mittal, Moorthy, and Bovik, 2012).

*2.2.2. Visual Scene Characteristics*

Even when considering the informational value of the number of pictures or videos in projects and their specific image characteristics such as color features, composition, and figure-ground relationships, we believe that there is further valuable information embedded within the images themselves. This additional information is likely to be relevant and can provide insights for predicting crowdfunding success. Specifically, we expand the taxonomy proposed in past research to add an additional category, visual scene characteristics, and we propose that capturing information about certain image attributes, such as the presence of faces, visual complexity, the number of identifiable concepts, and the ease of associating image content with existing schema, can contribute to our understanding of crowdfunding outcomes. To support our proposition, we draw upon the literature on visual scene processing, which sheds light on the significance of these image attributes.

*2.2.2.1. Number of faces.* The effect of facial images on consumer attention and behavior has been extensively studied in the advertising literature. Faces have been found to rapidly convey information and hold sociobiological significance (Dekowska, Kuniecki, and Jaskowski, 2008; Hershler and Hochstein, 2005). For instance, Guido et al. (2019) discovered that in print advertising, the presence of facial stimuli can direct consumer attention to images and make communications more noticeable in a cluttered environment. Consequently, images that include faces are more likely to attract attention. This hypothesis has found some support in research showing that the number of faces featured in GoFundMe campaigns is associated with the number of backers and the amount of money raised (Rhue and Robert, 2018). However, it is important to note that reflexive attentional responses to facial stimuli might not necessarily translate into behavioral responses, such as sponsoring crowdfunding projects (Guido et al., 2019). Zhang, Lyu, and Luo (2021) reported mixed findings regarding the impact of facial images on project funding, with results varying based on the type of project being featured. It appears that while facial images may increase attention, this alone may not be sufficient to drive crowdfunding success.

*2.2.2.2 Number of evoked concepts.* Scholarly work in imagery examines how consumers interact with aesthetic elements to frame category inferences, often occurs independently of their examination of the surrounding contexts, though category inferences also may be contingent on how people comprehend the scenes and scenarios in which they first interact with novel objects (Buetti et al., 2016; Torralba and Olivia, 1999). For example, in a scenario in which a tech-savvy entrepreneur develops an innovative new home juicer, representations might manifest as distinct superordinate categories such as "kitchen appliance," "human," and "fruit." Such categories offer only sparse information about what makes the juicer new or innovative. The categories offer structural regularity around a somewhat vague theme, but the theme is not random (Buetti et al., 2016; Torralba and Olivia, 1999); it relates some items in some meaningful ways to a target while excluding unrelated items (Buetti et al., 2016; Wolfe et al., 2003). A theme that incorporates the juicer may be a kitchen, but only if the objects in the scene are arranged to support the housing, preparation, and consumption of food. Processing these relationships is of paramount importance for planning and predicting suitable actions. We expect that multiple category cues influence consumers' inferences about products, and the more distinctly that images evoke concepts (i.e., concepts come to mind), the more they evoke functional and spatial relationships that help reduce ambiguity about what products are and do.

*2.2.2.3 Ease of concept identification.* By definition, "innovative products defy straightforward categorization" (Moreau et al., 2001, p. 496). As a result, it is not only the number of evoked categories that matters, but also the ease with which the categories come to mind. Blanchard, Aloise, and DeSarbo (2012) find a strong association between salience of information and the speed at which people recall it when making judgments; Moreau et al. (2001) find that the first knowledge structure that comes to mind is particularly important to a person's judgment of new products. Accordingly, we expect that the ease with which the most salient concept is identified helps make the elements of visual scenes in images displayed on crowdfunding platforms less ambiguous.

.

## 3. EMPIRICAL SETTING AND DATA

Kickstarter, a popular crowdfunding platform, has played a significant role in supporting projects, with over $5 billion in pledges from 18 million backers and funding for more than 190,000 projects (Kickstarter, 2020). The platform operates under an all-or-nothing funding model, where backers are only

charged if a project reaches its funding goal. Kickstarter has specific rules for technology projects, requiring them to showcase a non-rendered visual prototype to provide backers with a clear understanding of the project's development stages. These prototypes are displayed as title images on the browsing page. Photorealistic renderings such as technical drawings, computer-assisted drawings (CAD) models, or sketches are prohibited as they can potentially bias perceptions of the final product's appearance. The Kickstarter project database is regularly scraped and made publicly available by Web Robots.

In our research, we focus on the role of visual elements in informing potential backers about the functionality of new products. To narrow down our analysis, we specifically examine projects listed in the "technology" section of Kickstarter, excluding offerings classified in other functional categories such as comics and illustrations, food, music, games, and publishing. Our dataset comprises 16,439 projects, which includes both successful (4,187) and failed (12,252) projects for which we have complete data. Each project entry includes essential information such as project ID, funding goal, amount pledged, number of backers, whether it was featured as a staff pick, project launch date, and project deadline. Additionally, we collect the project's title, a brief text description, and a link to the project's title image visible in search results. To enrich our data, we augment the webrobots data by scraping full-text extractions of the project descriptions.

3.1. Baseline Variables

As baseline variable, we considered external factors that could influence the success of a project. One such factor is the potential feature of projects on Kickstarter's highly visible sections called "staff picks" or "projects we love," which has been associated with increased chances of success (Pisani, 2014). We also considered the geographic location of the project creator, as studies have shown that backers may have a higher willingness to pledge to projects based on the creator's location (Mollick, 2014). To account for seasonality and temporal effects, we included two control variables related to the project's launch date: the launch day of the year and the year of the project (ranging from 2009 to 2017). Additionally, we incorporated the duration of the project's funding window, measured in the number of days the project could accept funding. While Kickstarter initially allowed projects to be open for 90 days, they now encourage shorter durations, typically 30 or 60 days. Several project content characteristics that can influence project success. For example, projects with smaller funding goals are more likely to be backed and to achieve their objectives (Mollick, 2014). Accordingly, we controlled for funding goal size. Although many project creators were located outside the United States and used other currencies, the data scraper automatically converted the currencies using the exchange rate to USD at the time of scraping.

3.2. Text Variables

To assess the impact of text descriptions on potential backers, we analyzed both the short text description (blurb) and the full text description of the projects. For the full text analysis, we utilized the Linguistic Inquiry and Word Count (LIWC) dictionaries. Each dictionary (e.g., anger) that contain words that belong to different psychological, emotional, and linguistic dimensions (e.g., hate, mad, angry, frustr*). To use LIWC on a text, the text is compared with each dictionary and the percentage of words that match each category is calculated. We used LIWC-22 to create our 117 variables (see Boyd et al., 2022, for the list of dictionnaries). The LIWC-22 has been extensively validated using diverse sources of publicly available texts, such as social media posts, reviews, articles, stories, and conversations from platforms like Facebook, Reddit, Twitter, Yelp, and more. Compared to earlier versions like LIWC-2015, the LIWC-22 offers additional variables related to determiners, cognition, states, motives, affect, and social behaviors, making it more comprehensive and suitable for our study. Boyd et al. (2022) found slightly higher correlations among self-reports, judges, and LIWC using the LIWC-22, further validating its effectiveness.

For the analysis of the blurb, we employed six variables. To account for text descriptions that emphasized the project's innovativeness and superiority, we created two custom dictionaries containing the top 100 synonyms for the words "best" and "innovate," respectively. These variables aimed to capture textual claims of being innovative and superior. Additionally, we included four text summary variables generated by LIWC: analytical language, clout/persuasiveness, authenticity, and emotional tone. These variables have been commonly used in automated text extraction in the crowdfunding literature and have an influence on text persuasiveness (Pennebaker et al., 2014; Kacewicz et al., 2013; Newman et al., 2003; Cohn, Mehl, and Pennebaker, 2004).

3.3 Visual Data

Among the many ways to operationalize our image detail variables as measures, we opted to rely on measures that are both common in practice and readily accessible from common packages such OpenCV (a popular, free, open-source computer vision algorithms), the Python Imaging Library (PIL), or SciPy, with many variables adapted from Zhang and Luo (2023). Table 2 lists the measures.

For visual scene variables, we proposed three variables: the number of faces, the number of evoked concepts, and the ease of concept identification. For the number of faces, we use the number of detectable front upright faces based on the pre-trained Viola-Jones algorithm of Pech-Pacheco et al. (2020) which has been validated elsewhere. For the number of evoked concepts and ease of classification, there are numerous data sources and pretrained models for multilabel annotation to identify the concepts that humans who evaluate images are likely to evoke. We first chose to identify elements in images automatically, using Google's Cloud Vision API, which annotates input images with discrete labels that are then confidence-scored. When annotating images, the algorithm first relies on Google's Open Images data set (Krasin et al., 2017), which contains 478,000 crowdsourced images annotated by humans with more than 59 million image-level labels across 19,000 categories. Using TensorFlow (Google's internal machine learning framework), a deep convolutional neural network assigns multiple labels to the evaluated images. We therefore counted the number of labels extracted by the Google Vision API for which classification confidence was at least 50%.[5] We used maximum classification confidence across all identified concepts to measure ease of concept identification. Although there are alternative measures (e.g., average, median), capturing the ease of concept identification with the highest confidence label is consistent with past categorization research, which indicates that the most salient category exemplar presents distinct advantages for both recall (Blanchard et al., 2012) and inferences (Moreau, Markman, and Lehmann, 2001).

We argued the number of annotations provided by commercial APIs is a variable that is relevant to visual scenes because the annotations are likely to reflect concepts that are relevant to product usage. It would be surprising to find entrepreneurs in our Kickstarter data using depictions of concepts that are difficult to identify or have no relation to their innovative products, particularly in settings in which the entrepreneurs are attempting to communicate about their products to potential backers. Even if we were to assume that our arbitrarily chosen classifier could perfectly recover the ground truth labels that would be assessed by humans (which it does not), such visual elements may be unrelated to a product's usage scenario. Consider SHRU (see Figure 4), an intelligent cat toy offered on Kickstarter as "not another cat toy, it's a cat companion, design to be your cat's new best friend."

The toy looks like a small ball, the cat has whiskers and fur, and the image contains a caption. We seek to provide empirical evidence that our annotations capture aspects of the visual scene relevant to how the product is intended to be used. To do so, we randomly selected a subsample of 1,603 projects in our data sets and subjected them to a validation task using mTurk workers on Amazon. Specifically, we created a custom classification task in the mTurk API, whereby mTurk workers independently evaluated the labels identified by the Google Vision API for each project. Specifically, participants viewed the project title, product image, and "blurb" (short description).

---

[5] 50% is the default value for the vision API.

Table 2 – Image Details

Color features
- *Brightness*: the average value of the V channel in the HSV color space, normalized to [0,1].
- Saturation: the average value of the S channel in the HSV color space, normalized to [0,1].
- *Colorfulness*: a metric for how colorful the image is, computed as the sum of the standard deviation of the R-G and B-Y color channels, normalized to [0,1].
- *Contrast*: the standard deviation of the V channel in the HSV color space.
- *Warm Hue*: a metric for how much of the image's color is in the range of red, orange, and yellow, computed as the proportion of pixels whose H value in the HSV color space is within the range.
- *Clarity*: a metric for how clear or sharp the image is, computed as the proportion of pixels whose V value in the HSV color space is greater than or equal to 0.7.
- <u>Blur</u>: Degree to which shapes of surfaces are difficult to perceive, calculated as the variance of Laplacian of the grey-scaled image.
- <u>Image quality</u>: predicted human judgments based on the pre-trained LIBSVM by Mittal et al. (2012)

Composition
- *Diagonal dominance*: the average distance of a salient region's points to the two diagonals of an image and taking the minimum of those distances, which is then subtracted from zero.
- *Rule of third*: Using saliency maps from the StaticSaliencyFineGrained (OpenCV), adherence to the rule of third is calculate as the proximity of the centroid of the salient region and the nearest intersection in a 3x3 grid.
- *Vertical physical balance*: Proximity between the center of the image and the weighted center of the salient regions in the vertical axis based on saliency maps obtained from OpenCV.
- *Horizontal physical balance*: Proximity between the center of the image and the weighted center of the salient regions in the horizontal axis based on saliency maps obtained from OpenCV.
- *Vertical color balance*: Vertical color balance is a measure of how evenly the colors are distributed from top to bottom in an image, with values ranging from -1 (completely unbalanced) to 0 (perfectly balanced). This is computed using the superpixel pixel algorithm (Achanta et al. 2010)
- *Horizontal color balance*: Horizontal color balance is a measure of how evenly the colors are distributed from left to right in an image, with values ranging from -1 (completely unbalanced) to 0 (perfectly balanced). This is also computed using the superpixel algorithm (Achanta et al. 2010).
- <u>Number of segments</u>: Count of number of partitions (image segments) in an image, obtained using Otsu's binarization to identify the threshold that minimizes weighted within-class variance on a grayscale image, and count of number of edge changes (Gonzalez and Woods, 2002)

Figure-Ground Relationship
- *Size difference*: Using a GrabCut algorithm (Rother, Kolmogorov, Blake 2001), the image is segmented in foreground and background. The size difference is the difference in area size.
- *Color difference*: The Euclidian distance between the mean color values of the foreground and background regions provides the color difference.
- *Texture difference*: Using Canny edge detection, texture difference is the absolute difference between the proportion of edge pixels between the foreground and background regions.

Visual Scene Characteristics
- <u>Number of faces</u>: the number of detectable front upright faces based on the pre-trained Viola-Jones algorithm of Pech-Pacheco et al. (2020)
- <u>Number of evoked concepts</u>: Image-level annotations identified as having 50% confidence from the Open Images Dataset using the Google Vision API (Krasin et al. 2017) and Microsoft Azure
- <u>Ease of concept identification</u>: Classification confidence of the most confidently identified annotation detected from the Open Images Dataset using the Google Vision API (Krasin et al. 2017) and Microsoft Azure

**Note:** *Italics* identified variables operationalized as in Zhang and Luo (2023). Underlines refer to variables that we added.

**Figure 4. Sample Label Validation Task for SHRU, the Intelligent Cat Companion**

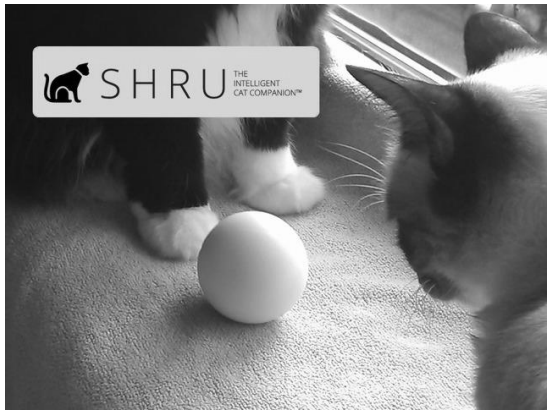

They then read:

> A computer program identified the following labels as part of the above image. Some of the labels relate to a broader scene that depicts how the product is meant to be used, and others have nothing to do with the scene being conveyed in the image. Which of the following labels are not related to how the product is meant to be used? Unselect the labels that do not directly relate to what you believe is the gist of the scene being conveyed in the image. [see instructions]"

Participants also answered two questions to help remove ambiguity about whether the annotations were relevant to the product: "To what extent do <u>all</u> the labels relate to one another to describe a product usage scenario?" (1, "not at all related to a scenario"; 9, "completely related to a scenario"), and then a nominal "yes" or "no" question about whether the image conveyed how someone would use this product.

The [see instructions] button displayed a layer that provided a sample task with labels accompanied by a training task to illustrate how to select relevant labels. Specifically, it showed the project title, image, and blurb for the SHRU as an example (see Web Appendix A). We then provided participants with further clarification related to relevance labels:

> Here we have a cat toy. Of all the labels, not all directly relate to how the toy is meant to be used. For instance, *cat*, *ball*, *play*, *small to medium sized cats*, *kitten* would all be obviously relevant to how the designer intended the product to be used (e.g., a ball that cats or kittens can play with). However, *carnivore* and *whiskers* (although both about a cat) do not relate to how the product is meant to be used. Likewise, *photo caption* isn't helpful in explaining the use of the object. We would expect you to <u>unselect</u> *photo caption*, *carnivore*, and *whiskers*.

The mTurk task was structured such that each project had five workers perform the evaluation. The task was completed using 403 unique workers who evaluated 1,603 images for total compensation of $1,141.80. We classified any label as validated if at least 3 of the 5 workers who performed the task agreed that the label was related to the product usage. We assumed the label validation rate would be rather high, because it would be surprising to find the concepts unrelated, particularly regarding the most salient object in the scene (i.e., most easily identifiable), and because they appeared in a setting in which entrepreneurs were attempting to communicate a product to potential backers.

As expected, on average, most labels (84%) captured by the Google Vision API were explicitly retained as related to the product once validated by the mTurk workers, with an average reduction of 1.97 labels. The number of original labels had a .938 ($p < .001$) correlation with the number of validated labels, which reflects the idea that a rational entrepreneur would not contrive an unrelated scene when accurate communication is critical. The average decrease in primary classification confidence was 1.31%, which correlated with the unvalidated primary classification confidence of .891 ($p < .001$). These label validation results substantiate our assumption that most of the distinct labels cataloged by the visual API consist of objects that relate spatially, or categorically to what we described conceptually as the product.

## 4. PREDICTIVE VALUE OF VISUAL INFORMATION

To this point, we have documented that visual information is likely to play a role in crowdfunding success, and we have presented a series of concepts: visual counts and image details. In this section, we use our empirical data set from Kickstarter to illustrate how such information can be automatically extracted and used within machine-learning models to show the importance of visual information.

4.1 Forecasting Framework

To assess the importance of visual information contained in images, we compare five sets of variables: (1) baseline (six project characteristics only), (2) baseline + textual details (from blurb and main text; 123 extracted variables) (3) baseline + visual counts (number of images and videos; two variables), (4) baseline + image details, and (5) all variables. We fit the same models for the five sets of variables.

4.2 Comparative models

Our outcome is (logged) total dollars raised $y_i$ and predictors $x_i = \{x_1^p, x_2^p, \ldots, x_N^p\}$ for $N$ crowdfunding projects. Our statements regarding visual information are agnostic to the kind of classification model used. We needed to choose a learning model that can accommodate unexpected non-linearities and variable selection to make sure that we truly capture the predictive value of visual information.

*4.2.1 Regularized Regressions (Ridge and Lasso).* First, we considered regularized regression given large predictor spaces and potential for overfitting. Regularized regression methods, such as ridge and lasso, are popular techniques to conduct regressions while avoiding overfitting and improve model performance on out-of-sample data. To do so, they add a regularization term to the regression objective function. In both methods, the objective is to minimize the sum of squared errors between the predicted outcome and the actual outcome, subject to a constraint that is a function of the predictor variables.

Ridge regression adds a penalty term to the squared error loss function, which is proportional to the sum of the squared values of the regression coefficients. The loss function minimized in Ridge is given by:

$$Loss_{ridge} = \sum_{i=1}^{N} L(y_i, \hat{y}_i) + \lambda \sum_{j=1}^{p} \beta_j^2 \tag{1}$$

where $\lambda$ is the tuning parameter that controls the amount of shrinkage applied to the coefficients and $L(y_i, \hat{y}_i)$ is the squared error loss. The shrinkage parameter $\lambda$ penalizes the sum of squared coefficients in the model and shrinks them towards zero. The larger the value of $\lambda$, the more the coefficients are shrunk towards zero, resulting in a more biased but less variable model.

Lasso regression, instead of using a shrinkage parameter affect coefficients, imposes a penalty to the objective function of a linear regression based on the $L_1$ norm of the vector of coefficients (i.e., sum of absolute value of the coefficients) which is weighted against the residual sum of squares between the predicted values and observed values using a regularization parameter $\lambda$:

$$Loss_{lasso} = \sum_{i=1}^{N} L(y_i, \hat{y}_i) + \lambda \sum_{j=1}^{p} |\beta_j| \tag{2}$$

This has the advantage of providing a simpler and more interpretable model, by selecting only the most relevant predictor variables and encouraging some to be exactly zero.

*4.2.2 Bayesian Additive Trees.* Bayesian Additive Regression Trees (Chipman, George, and McCulloch, 2010) have been shown to provide a flexible approach to make predictions while avoiding strong parametric assumptions (Hill, Linero and Murray 2020), particularly when cross-validation is used to choose hyperparameters (Dorie et al. 2019). Assuming that $(y_i, x_i)$ are i.i.d., a single tree in a BART models the expected value of the continuous outcome as:

$$E(Y|x) = g(x; T, M), \tag{3}$$

where $g(x; T, M)$ is the function that follows the rules described by regression tree $T$ and $M$ is the set of terminal node predictions for each node. A single tree is built via recursive partitioning, which divides up the $p$-dimensional space of the predictors into mutually exclusive areas. Each area is as homogeneous as possible (i.e., contain observations of similar expected value). In addition to building different algorithms to estimate single, researchers have worked on ensemble methods that produce their forecast based on more than one decision tree. This includes boosted trees (e.g., AdaBoost) and bootstrap aggregated trees (e.g., random forests). Such ensembles typically outperform single trees due to their ability to avoid overfitting (Hastie, Tibshirani, and Friedman, 2008).

One notable advancement in ensemble methods for decision trees has been the development of Bayesian Additive Regression Trees (Chipman, George, and McCulloch, 2010). Still assuming that $(y_i, x_i)$ are i.i.d., Bayesian Additive Regression Trees model predict the outcome as follows:

$$E(Y|x) = \Phi \left[ \sum_{j=1}^{m} g(x; T_j, M_j) \right], \tag{4}$$

where $g(x; T_j, M_j)$ is still the function that allows one to follow the tree-specific regression rules described by the tree $T_j$ and $M_j$ is the set of terminal node predictions (i.e., expected outcome values) for each node. For estimation, Chipman, George, and McCulloch (2010) proposed a Bayesian backfitting MCMC algorithm that incorporates a regularization prior to avoiding having a single tree too influential and can outperform many existing machine learning methods on prediction tasks. It has also been used as a predictive benchmark in marketing (Schoenmueller, Blanchard and Johar 2022).

*4.2.3 XGBoost.* Another ensemble approach, eXtreme Gradient Boosting (XGBoost; Chen and Guestrin 2016), is popular in marketing (e.g., Rafieian and Yoganarasimhan 2021) and operations management (Schaer, Kourentzes and Fildes 2022). Similar to BART, XGBoost is an ensemble method that combines multiple decision trees to make predictions. However, the approach is different. XGBoost seeks to find the optimal set of decision trees by minimizing a following loss function:

$$Loss_{XGBoost} = \sum_{i=1}^{N} L(y_i, \hat{y}_i) + \sum_{j=1}^{m} \Omega(T_j), \tag{5}$$

where the first term represents the data loss, and the second term $\Omega(T_j)$ is a regularization term to penalize large values in the terminal node predictions and trees. XGBoost uses gradient boosting to build each tree by iteratively correcting the errors made by previous trees. From an initial set of parameters, it calculates the negative gradient of the loss function as pseudo-residuals. The tree is then fitted to these pseudo-residuals, and this process continues with a weighting factor determined by a learning rate parameter. XGBoost also includes techniques like shrinkage, column subsampling, and early stopping to improve performance and prevent overfitting.

BARTs and XGBoost differ in several ways. On the one hand, XGBoost incorporates several regularization techniques and frequently achieves the highest performance in addition to being scalable for large datasets and high-dimensionality. On the other hand, BARTs provides several advantages when it comes to interpretability: its Bayesian approach allows one to quantify uncertainty, impute missing values and intervals using posterior distributions. This can be particularly helpful when trying to assess for the relative importance of subsets of features or to do significance tests. It is also considered to be simpler to implement because packages require fewer tuning parameters.

4.3 Predictive Results

We divided the data into an in-sample training step and an out-of-sample testing set to train and evaluate our models. We used the training set (80%) to train the model and the testing set (20%) to evaluate it. For regularized regressions, we first scaled the variables prior to using cross-validation on mean-square errors within the training data to select the tuning parameter $\lambda$ for each of Ridge and Lasso. For BARTs, we used 10-fold cross valuation within the training set to choose the hyperparameters in our classifier (e.g., number of trees $m$, shrinkage parameter $k$) as implemented in BARTMachine (Kapelner and Bleich 2013). For XGBoost, we also used 10-fold cross-validation as it has many parameters (eta, max depth, subsample, column sample by tree, number of rounds and early stopping rounds). The decision of using cross-validation across all models was done to minimize our use of researcher degrees of freedom (i.e., our choice of hyperparameters being responsible for model results). Table 3 shows the results for in-sample RMSE (root mean squared errors) and MAE (mean absolute errors). We note that the RMSE and MAE for Ridge and Lasso are based on final descaled predictions so that the values are comparable to that of BART and XGBoost and so that they are more easily interpretable. We also note that for missing values for some image characteristics which could not be computed, we used the KNNimpute function in the R library Caret for Lasso, Ridge and XGBoost whereas we used the recommended procedure by Kapelner and Bleich (2013; see Kapelner and Bleich 2014)

Some results are consistent across all four models. Compared to a model trained on only the baseline variables, the addition of variables (whether they are counts of visuals, text, or image details) improves both in-sample and out-of-sample RMSE and MAE metrics. Moreover, the addition of only visual count variables (number of pictures and number of videos) has a strong impact, nearly as much as the addition of the entire LIWC dictionary on the full text and selected measures extracted from the blurbs. When using BART or XGBoost, we find that the addition of these two variables reduces RMSE by approximately 20%, which is a bit more than adding all textual variables and more than adding all image details. Additionally, we find that adding textual details seems to have more incremental impact than images - even if images help beyond the baseline variables. Finally, we note that including all variables provides incremental predictive value and that the model with all variables predicts best in out-of-sample results across all models.

We also find some differences between the models. It is notable that BART and XGBoost fits better than Ridge or Lasso across all variable sets, in and out of sample. Their ability to improve predictive accuracy is most notable when few variables are used, but the improvements are also notable when the entire set of variables is used for out-of-sample RMSE (XGBoost=2.35; BART=2.35; Ridge=2.54; Lasso=2.54) and MAE, which is not optimized directly by any of the models (BART=1.88; XGBoost=1.88; Ridge=2.07; Lasso=2.07).

**Table 3. Kickstarter: Comparisons of Prediction across Subsets of Variables**

| Model | Variable sets | In-sample RMSE | In-sample MAE | Out-of-sample RMSE | Out-of-sample MAE |
|---|---|---|---|---|---|
| Ridge | 1) Base | 3.6232 | 2.8988 | 3.4593 | 2.9123 |
| | 2) Base + Visual counts | 3.3868 | 2.3885 | 2.8455 | 2.3722 |
| | 3) Base + Text details | 3.2502 | 2.1387 | 2.7277 | 2.2185 |
| | 4) Base + Image details | 3.5279 | 2.7246 | 3.2850 | 2.7300 |
| | 5) All | 3.1482 | 2.0118 | 2.5426 | 2.0734 |
| Lasso | 1) Base | 3.4899 | 2.8982 | 3.4597 | 2.9137 |
| | 2) Base + Visual counts | 3.0079 | 2.3851 | 2.8423 | 2.3685 |
| | 3) Base + Text details | 2.8435 | 2.1151 | 2.7308 | 2.2108 |
| | 4) Base + Image details | 3.3583 | 2.7246 | 3.2913 | 2.7343 |
| | 5) All | 2.7010 | 1.9957 | 2.5408 | 2.0663 |
| BART | 1) Base | 3.1243 | 2.5784 | 3.1893 | 2.6316 |
| | 2) Base + Visual counts | 2.4555 | 1.9638 | 2.4874 | 1.9946 |
| | 3) Base + Text details | 2.3648 | 1.8891 | 2.5316 | 2.0395 |
| | 4) Base + Image details | 2.9235 | 2.4034 | 3.0468 | 2.4972 |
| | 5) All | 2.2371 | 1.7817 | 2.3542 | **1.8838** |
| XGBoost | 1) Base | 3.1335 | 2.5902 | 3.2050 | 2.6494 |
| | 2) Base + Visual counts | 2.4494 | 1.9605 | 2.4849 | 1.9938 |
| | 3) Base + Text details | 2.3421 | 1.8756 | 2.5461 | 2.0523 |
| | 4) Base + Image details | 2.9422 | 2.4185 | 3.1074 | 2.5454 |
| | 5) All | **2.2039** | **1.7551** | **2.3521** | 1.8864 |

Overall, the results show that the addition of visual information, particularly visual count variables, improves the performance of all three models on in-sample and out-of-sample metrics. Additionally, the BART and XGBoost model outperforms the Ridge and Lasso models across all variable sets, likely indicative of their ability to better accommodate non-linearities using the raw data (i.e., our implementations of Lasso and Ridge did not include two-way interactions). Across model, we also find that although the image details variables reduce out-of-sample performance beyond the base model, their predictive value is notably lower than the two visual count variables. Visual count variables seem to help predict even better than all text details, when BARTs and XGBoost are given. Given the similar performance of BART to XGBoost and the usefulness of being able to sample the posterior for uncertainty quantification as discussed below, the subsequent descriptive analyses will focus on the BART model results.

4.4. Importance of Visual Counts and Image Details

We have argued that visual elements are important predictors and should be incorporated when predicting crowdfunding success; visual information provides significant predictive power in both in-sample and out-of-sample. Although these findings support that adding visual count variables and image details significantly improves prediction, there is no evidence that removing image variables would significantly affect fit. One way to obtain such evidence is to retrain a BART model without any specific covariate and compare the pseudo-R2 of the model with that of BART models retrained using "null" samples where the predictors are removed (see Bleich et al. 2014). Using 100 resamples where we remove all image details (but kept visual counts), we find that all of the resamples see a reduction in pseudo R2. If we remove only the new image details we proposed (visual scene elements, with the addition of blur, quality/noise and number of segments), we find that 85% of resamples saw a reduction in pseudo-R2. This suggests that not only are image details important beyond visual count variables but that the proposed predictors can add important information for prediction.

Next, we use two strategies from interpretable machine learning (e.g., Molnar 2020; Molnar, Casalicchio and Bischl 2018) to provide confidence in the importance of the visual attributes: feature importance and partial dependence plots.

*4.4.1 Feature importance*

According to Molnar (2020), feature importance in machine learning refers to the increase in the prediction error of the model that occurs when we break the relationship between the feature and the true outcome. There exist many ways to calculate variable importance scores. In regression, one often compares standardized regression coefficients as differences in size of unstandardized coefficient could be due to feature scaling. In BART models, variable importance is commonly assessed via inclusion proportions in the post burn-in MCMC iterations (Bleich et al. 2014). Here, we use the proportion of times each variable is chosen in a tree among all the posterior draws of the sum-of-tree model. In Figure 3, for the sake of illustration, we present the Top 40 Highest Feature Importance features along with the confidence intervals (95%). In Table 4, we present some summary statistics coded by variable type.

**Figure 3. Kickstarter: Top 40 Highest Variable Importance (% of trees including the variable)**

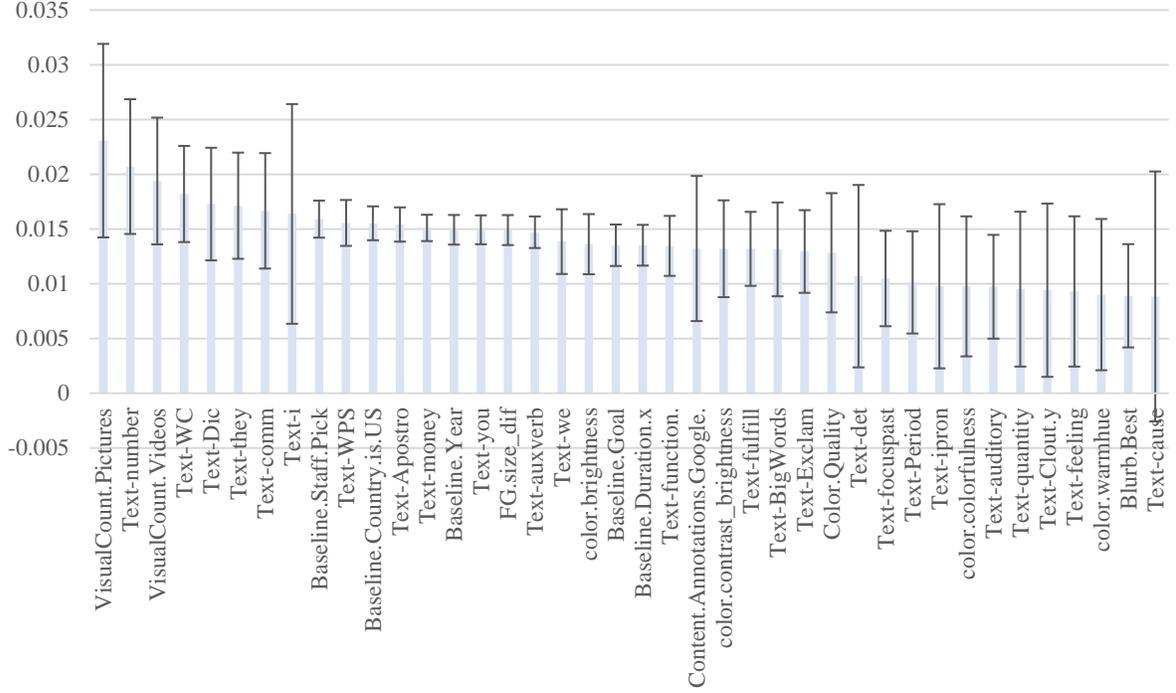

**Table 4. Kickstarter: Summary Statistics on Feature Importance (% trees) by Category**

|  | Number of variables | Prop. Inclusion | Fraction Sig. |
|---|---:|---:|---:|
| Visual: Count | 2 | 2.12% | 100% |
| Baseline | 6 | 1.27% | 83% |
| Visual: Figure-Ground | 3 | 0.82% | 33% |
| Visual: Color | 7 | 0.75% | 100% |
| Text-Description | 117 | 0.63% | 62% |
| Visual: Scene | 6 | 0.52% | 83% |
| Text-Blurb | 6 | 0.32% | 50% |
| Visual: Composition | 8 | 0.24% | 63% |

Consistent with the results in Section 3, we find that visual count variables are important. Whereas the average variable has a .63% probability of being included in the final tree, the visual count variables (number of pictures: 2.31%, number of videos: 1.94%) is higher than most. Moreover, several of the image detail measures are most likely to be included. All color image detail variables have a significant probability of inclusion, with brightness (1.36%, SE=.27%), contrast-brightness (1.36%, SE=.27%). For figure-ground, the size difference is likely to be included (1.49%, SE=1.32%) however texture-difference and color difference are not. For composition, horizontal color balance (0.83%, SD=.69%), number of segments (.38%, SE=.37%) are most important.

Focusing on the variables that are novel in our research (i.e., beyond Zhang and Luo 2023), we find that the variables that are novel to our inquiry for color (blur: .46%, SE=.43%; quality-noise: 1.28%,

SE=.54%) and composition (number of segments) are significant. For our visual scene variables (all novel), the most important is the number of annotations provided by Google (1.32%, SD=.66%) but the maximum classification confidence is not (.27%, SE=.29%). It also seems that importance of the number of annotations from Google Vision is greater than that provided by Microsoft Azure (.09%, SE=.02%), and that the maximum classification confidence provided by Azure (.09%, SE=.02%) has a non-zero probability of inclusion. Finally, we note that the number of upright faces has a small but significant probability of inclusion (.09%, SE=.05%).

Looking at whether confidence intervals exclude zero across features allows us to look at which type of features are more likely to have non-zero inclusion. We find that the confidence interval excludes zero for all visual count and base variables other than day of year, and nearly two thirds of the textual variables be that from the blurb and text have a significant probability of being included in the model. Of the image details, all color variables (brightness, contrast, blur, quality/noise, colorfulness) are significant and they are, on average, most important. Only one of the figure-ground variables had a non-zero probability of inclusion (size difference: 1.49%, SE=.14%). The visual scene variables all had significant probability of inclusion, other than the maximum classification confidence provided by Google Vision.

Although these findings support that adding visual count variables and image details improves prediction, there is no evidence that removing image variables would significantly affect fit. One way to obtain such evidence is to retrain a BART model without any specific covariate and compare the pseudo-R2 of the model with that of BART models retrained using "null" samples where the predictors are removed (see Bleich et al. 2014). Using 100 resamples where we remove all image details (but kept visual counts), we find that all of the resamples see a reduction in pseudo R2. If we remove only the new image details we proposed (visual scene elements, with the addition of blur, quality/noise and number of segments), we find that 85% of resamples saw a reduction in pseudo-R2. This suggests that not only are image details important beyond visual count variables but that the proposed predictors can add important information for prediction.

*4.4.4 Exploring functional forms - Partial Dependence Plots*

All the prior analyses have focused on illustrating whether the features are useful for prediction, and have been agnostic to how any features associate with the outcome. Yet, one of the most useful aspect of tree-based models is that they can flexibility accommodate non-linear relationships between covariates and outcome. One way to explore the function form of the relationship between any single feature and the outcome is to use partial dependence functions (PDFs; Friedman 2001). Partial dependence functions (PDFs) which provide the expected value of the response variable as a function of a particular predictor variable, while averaging out the effects of all other predictor variables in the model. For any given predictor variable, one first fixes the values of all other predictor variables, and then computes the expected value of the response variable over a range of values for the chosen predictor variable. This process is repeated for different reference values of the other predictor variables, and the resulting partial dependence functions are averaged to obtain an overall estimate of the relationship between the response variable and the chosen predictor variable. Plotting the results of this process through Partial Dependence Plots can be useful to identify regions of the predictor variable space where the response variable is particularly sensitive to changes in the predictor variable, or to visualize the shape of the relationship between the response variable and the predictor variable. To the extent that interpretable features are used in the model, this approach provides interpretability to otherwise opaque models (c.f., Molnar 2020).

In Figure 4, we present plots for several of the proposed visual variables that had the highest proportion inclusion. Nevertheless, we notice an intuitive pattern for the relationship between variables and log dollars raised. For example, there is a significance increase in log dollars raised going from zero to one video, but less evidence going from one to more which is content with the recommendation that authors performing empirical analyses of Kickstarter data should at least include a control for whether the project has at least one video. Intuitively, we find that consideration of image quality (i.e., absence of noise) is fairly linearly associated with dollars raise but blur is not. For images, the number of pictures is generally positively associated with dollars raised, there's slowly diminishing marginal returns for doing so beyond five. Moreover, the number of concepts seems to peak at five, consistent without our intuition that the elements contained in the image might help disambiguate what the product is and what it does without introducing unnecessary complexity. The reduction in dollars raised associated with having too many associations may be related to the findings on number of image segments (a composition feature) that more annotations also likely capture more image segments.

4.5 Robustness: Predictive value of number of pictures: the role of due to post-failure project editing

We have found that visual count variables (number of images and number of videos) are important predictor of success (log dollars raised). This association is unlikely to be causal as numerous factors can simultaneously predict the number of visuals and odds of success (stage of prototype development). We cannot rule out that possibility. However, it is important to note that projects can be edited once the campaign is finished. Anecdotally, we have noticed that some projects have removed images and edited the text post failure – presumably in an attempt to defend against copy-cats and to redirect potential customers outside of the platform[6].

---

[6] Some projects edited the title to indicate that the project was cancelled (e.g., "XYZ (Cancelled)") whereas others edited the text directly (e.g., "The great news is that even though the Kickstarter campaign is ending, the project will continue! You can follow our progress at XYZ.com.")

**Figure 4. Partial dependence plots for added image characteristics (Y-axis is partial effect)**

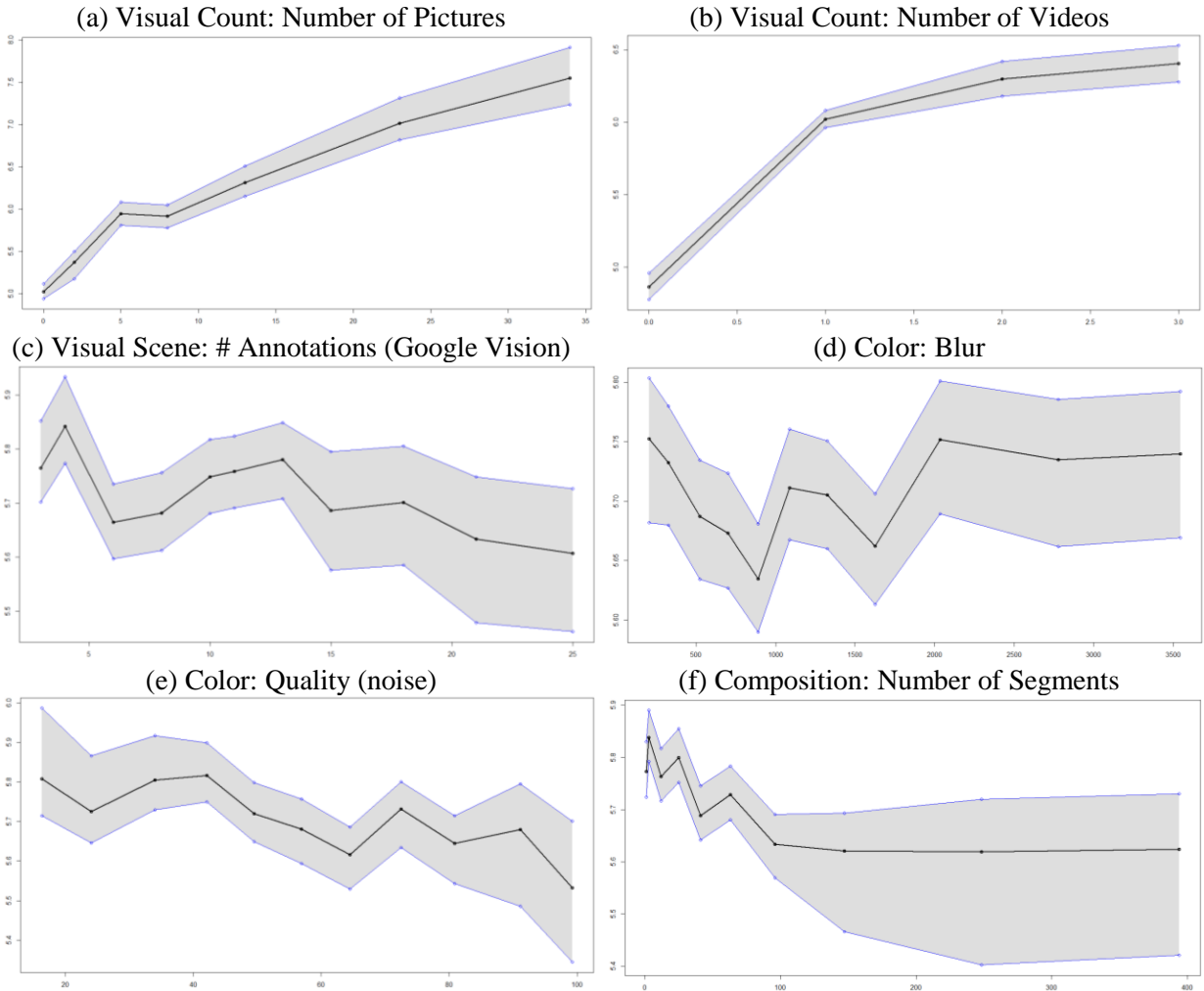

(a) Visual Count: Number of Pictures
(b) Visual Count: Number of Videos
(c) Visual Scene: # Annotations (Google Vision)
(d) Color: Blur
(e) Color: Quality (noise)
(f) Composition: Number of Segments

Note: plotted at quintiles.

If that's true, then the association we find between number images and dollars may be due to unsuccessful projects removing their images. To investigate whether that is a likely explanation, we created an interface to evaluate blinded project descriptions to see if the text of failed projects (including cancelled and suspended) with less than 1000 characters in the project description had been edited post failure. This interface did not highlight anything else about the project other than its description, so the coded was blind to the number of images and videos for the project. Out of 3046 projects manually inspected, 146, or 4.79%, had clear modified their listing after the fact.

Comparing the projects (modified versus not) on project characteristics, we find a few differences. First, failed projects modified after the fact had fewer pictures on average (0.27 vs. 1.14, p<.01) and videos (.13 vs. .41, p<.01). This suggests that, indeed, failed projects who edited their listing also scrubbed their pictures and videos. Second, we found that these edited projects were also marginally *more* likely to have been labeled staff picks by Kickstarter (3% vs 1%, p=.06), and had raised more money (log dollars of 5.05 vs. 2.45) despite having failed by Kickstarter standards and not received funding. This would suggest that the importance of the number of pictures and number videos depends on whether it is indicative of pre-raise video, at least to an extent.

4.6. Summary

We investigated the importance of visual count variables (number of pictures and videos) and image details. The results suggest that visual information has considerable predictive value. Specifically, although summarizing visual information into counts (the number of pictures and videos) is most helpful, the title image of the project contains important additional information for predicting success. Easily extracted image details pertaining to color, composition and figure-ground and importantly, and we have provided evidence that novel automatically extracted visual scene features such as the number of upright faces, and the number of concepts and classification confidence are helpful.

## 5. DISCUSSION AND CONCLUSION

The fields of operations management, marketing, and management have a rich history of using technological tools to enhance and expedite processes related to new idea development. Within this context, there has been a focus on identifying novel predictors of crowdfunding success and utilizing prelaunch processes and tools to aid entrepreneurs in selecting the most promising ideas. In line with this trend, we explore the untapped potential of automated extraction of visual information as a valuable data source for studying crowdfunding in the domains of marketing, management, and operations management.

To begin, we conducted a comprehensive review of the literature to identify existing visual information features that can be extracted from project listings, including visual counts such as the number of pictures and videos, as well as image details encompassing color, composition, and figure-ground relationships. Building upon these established features, we proposed three additional variables (blur, quality/noise, and number of segments) that we believed to be promising in predicting crowdfunding success. Furthermore, drawing insights from the literature on visual scene dynamics, we advocated for the inclusion of visual scene elements, such as the number of faces, the number of evoked concepts, and classification confidence.In our empirical analysis, focused on Kickstarter projects, we employed nested comparisons of baseline, text, and visual features across various predictive models (lasso, ridge, BART, and XGBoost) to demonstrate the incremental predictive value of visual features as a whole. Additionally, using techniques from interpretable machine learning and our trained BART model, we highlighted the significance of individual features through proportion inclusions and covariate importance tests. We also utilized partial dependence plots to illustrate potential non-linear relationships. Through our research, we provide several valuable contributions to both researchers and practitioners.

5.1. Research Implications

Although empirical research on crowdfunding generally has considered visual information images by counting the number of pictures and videos attached to the project page (if at all; see Table 1) or focused on manually coded visual features such as founders' race (Younkin & Kuppuswamy 2018), we build on recently literature in marketing that has shown the usefulness of automatically extracting visual attributes can help predict firm success (Zhang and Luo 2023; Zhang et al. 2022) to show that automatically extractable visual features can predict crowdfunding success. In doing so, we expand not only the list of documented important visual attributes across established categories of color, composition and figure-ground by adding blur, number of segments, and image quality (noise) but also provide evidence of the usefulness of adding visual scene elements such as the evoked concepts and faces. Although we found that visual scene elements do incrementally predict crowdfunding success, we did find that image details contributed less to predictive accuracy than visual count variables. One possibility is that we did not use the best available APIs to identify our constructs. Another is that we only focused on the title image, and extracting image details from all visuals would be more beneficial. We encourage future research to look into how to summarize the information image details across multiple images.

Although the study of visual scene dynamics is not a novel area of research, the study of how products are visually processed typically has been assessed through individual human judgments and based on experimental data (e.g., Noseworthy and Trudel, 2011; Veryzer and Hutchinson, 1998). This individual judgment process is resource-intensive and time-sensitive, and it necessitates a trade-off of ecological validity for internal validity. Therefore, and noting the impracticality of using human judges to classify tens of thousands of project images according to whether the visual elements are relevant to a crowdfunding product, we advocate for the use of validated commercial APIs to extract the number of annotations and classification confidence, as proxies for number of concepts and ease of concept identification. We provided preliminary evidence that visual scene elements can inform crowdfunding success, and validated this with a separate task of human judgments of relevancy.

5.2. Practical Implications

For consumers, it is difficult to predict crowdfunding success, because many unobservable variables (e.g., characteristics of the product creator, product complexity) simultaneously influence product creators' strategies and outcomes. Rather than developing a statistical model or machine learning algorithm that would outperform the prediction of crowdfunding success—a task many authors have attempted (e.g., Etter, Grossglauser, and Thiran, 2013; Greenberg et al., 2013)—we used our knowledge of visual information processing and visual scene dynamics to identify potentially important predictors. We hope our efforts motivate researchers who are interested in developing predictive models to carry out the simple task of augmenting their data sets with the easily extracted visual counts and image details. Tools made available to non-technical investors (e.g., Sidekick.epfl.ch, Kicktraq) generally focus on textual and baseline characteristics and we hope that our findings about the important of visual counts and image details can help them make scalable products that can better predict success.

For project creators, product designers and marketers are responsible for both the selection and composition of the focal image and various other factors that could also influence funding outcomes. Whereas Kickstarter and platforms tend to offer advice that is meant to be applicable to a broad range of project, the advice given is general.[7] For example, Kickstarter recommends keeping things simple and this aligns well with our observation that too many image segments and evoked concepts (annotations) is associated with lower log dollars raised. Likewise, it mentions using high resolution assets which correlates with our finding that image quality (i.e., lack of noise) is important. However, a pre-trained model that natively handles non-linearities and interactions, such as ours, can help companies select alternative pictures to feature as the splash image. From a set of baseline characteristics and baseline variables that are project specific, one can feed in alternative pictures and see how odds of success are impacted. We note that although such predictive exercises are helpful, they would still be unlikely to be adequate substitutes for A/B testing that could be implemented on product pages.

For platforms like Kickstarter, the success of campaigns directly impacts their revenue as Kickstarter only distributes funds and earns it commission if the project exceeds its goals. It then becomes essential for such platforms to identify projects that are likely to succeed and find ways to understand funder dynamics as a function of marketing tools available (Kim et al. 2020). Platforms have different tools at their disposal, some more indirect (e.g., altering the duration window) and some more direct (e.g., highlighting projects, altering fees). One of the most commonly discussed levers at the disposal of the platform is the labeling and highlighting of projects as "staff picks" or "projects we love" which have been liked to increased odds of success (Pisani 2014). Yet, platforms are not able to highlight every single project and projects only generate funds for the platform is the project exceeds the stated goals. One possibility is that the platform strategically allocates staff pick labels not to provides that are clearly going to exceed their funding thresholds, but instead focus on giving staff picks to project that are likely to be completing at levels of funding near the campaign's goal and where the staff pick label can push them across the finish line. Although the staff pick feature is associated with increased (log) dollars when looked at descriptively (M=5.20 without; M=10.53 with; p<.01) and through BART covariate importance tests (p<.01), it is unclear whether the difference in dollars raised is due to an effect of highlighting the product via staff picks or whether it is due to selection of projects that are, at least, highly unlikely to fail.

---

[7] https://www.kickstarter.com/help/images

5.3. Summary

In this article, we have built on a recent body of work in visual processing and marketing that has studied how interpretable image details and attributes can be automatically extracted to predict firm-level outcomes (Zhang and Luo 2023; Zhang et al. 2022). One could wonder whether such efforts are necessary when modern machine learning algorithms have the capability to automatically learn patterns and feature representations from data. In our view, such feature engineering remains crucial for achieving accurate predictions and eventual theory development. First, we have highlighted that visual count information (number of pictures, number of videos), a feature that is not extracted from image details, is the most important variable in predicting the outcome. Second, relying solely non-interpretable models (e.g., CNNs) for prediction can fall short in practice unless one has an immense amount of data as automatic feature extraction can introduce unnecessary complexity, diminish performance, and make interpretation more challenging. Third, past research in marketing has demonstrated the significant impact of feature engineering on seemingly visual data. For instance, Blanchard, Dyachenko, and Kettle (2020) showed that feature engineering enabled a simple Bayesian model to outperform a convolutional neural network (CNN) that did not incorporate researcher-extracted features. This highlights the potential value of carefully engineered features even in visual domains. Ultimately, we aim to emphasize the importance of including relevant features in predictive models. By showcasing the predictive power of these feature sets, we hope to not only encourage researchers interested in empirical analyses to consider and incorporate them into their own analyses, but also hope to stimulate research that highlights novel data sources such as social media (Cui et al. 2018), tags (Nam and Kannan 2014), and textual features (Toubia, Berger and Eliashberg 2021) and facilitates the generation of important features.

**Web Appendix A - mTurk Image Classification Validation**

Below you will find a product, a product description, and an image.
**${Title}**
${Blurb}
${image}
A computer program identified the following labels as part of the above image. Some may be relevant descriptions of the image but have nothing to do with the product.
Which of the following labels are about the product? Unselect the labels that are not about the product in the description. [see instructions]
${cboxes}

**[begin instructions menu]**
In these tasks, we want you to help you identify which **labels are related to the product in the image**. For example, in the SHRU:

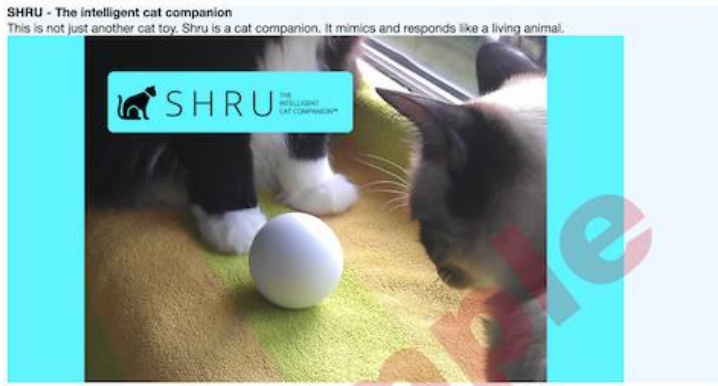

we have a cat toy. Of all the labels, not all are directly relevant to the product. For instance, *cat*, *ball*, *small to medium sized cats*, *kitten* would all be obviously relevant to the product. *Play* is an important part of the product concept, but *carnivore* and *whiskers* (although both about a cat) may not have much to do with the product. Likewise, *photo caption* isn't helpful. We would expect you to **unselect** *photo caption*, *carnivore,* and *whiskers*.
**[end instructions menu]**